\documentclass{article}

\usepackage{microtype}
\usepackage{graphicx}
\usepackage{subcaption}
\usepackage{booktabs} 
\usepackage[version=4]{mhchem} 
\usepackage{multirow} 
\usepackage{tabularx}
\usepackage{xcolor}
\usepackage{array}
\usepackage{algorithmic}

\newcommand{\g}[1]{\textcolor[gray]{0.6}{#1}}

\renewcommand{\algorithmiccomment}[1]{\hfill $\triangleright$ #1}
\newcommand{\codecell}[1]{{\normalfont\ttfamily\upshape #1}}

\usepackage{hyperref}

\usepackage[accepted]{icml2026}

\usepackage{amsmath}
\newcommand{\RETURN}{\STATE \textbf{return} }
\usepackage{amssymb}
\usepackage{mathtools}
\usepackage{amsthm}

\usepackage{threeparttable}

\usepackage[capitalize,noabbrev]{cleveref}

\theoremstyle{plain}

\theoremstyle{definition}

\theoremstyle{remark}

\usepackage[textsize=tiny]{todonotes}

\icmltitlerunning{Beyond Adam: SOAP and Muon for Faster, Label-Efficient Training of Machine Learning Interatomic Potentials}

\begin{document}

\twocolumn[
  \icmltitle{Beyond Adam: SOAP and Muon for Faster, Label-Efficient Training of\\Machine Learning Interatomic Potentials}




  \icmlsetsymbol{equal}{*}

  \begin{icmlauthorlist}
    \icmlauthor{Gil Harari}{equal,seas}
    \icmlauthor{Yoel Zimmermann}{equal,seas}
    \icmlauthor{Ola Tangen Kulseng}{seas}
    \icmlauthor{Laura Zichi}{seas}
    \icmlauthor{Chuin Wei Tan}{seas}
    \icmlauthor{Marc L. Descoteaux}{seas}
    \icmlauthor{Boris Kozinsky}{seas,bosch}
  \end{icmlauthorlist}

  \icmlaffiliation{seas}{John A. Paulson School of Engineering and Applied Sciences, Harvard University, Cambridge, MA, USA}
  \icmlaffiliation{bosch}{Robert Bosch LLC Research and Technology Center, Watertown, MA, USA}

  \icmlcorrespondingauthor{Boris Kozinsky}{bkoz@seas.harvard.edu}

  \icmlkeywords{Machine Learning, ICML}

  \vskip 0.3in
]



\printAffiliationsAndNotice{\icmlEqualContribution}  

\begin{abstract}
Machine learning interatomic potentials (MLIPs) have become a hallmark of AI for scientific simulation. 
While efforts on new architectures and datasets have led to increasingly accurate and general models, the choice of optimizer for training has largely remained unexplored, defaulting to Adam and its variants in the community.
Here, we implement and systematically compare a class of recently proposed matrix-structured optimizers, including Muon, SOAP, and the hybrid SOAP-Muon, for training NequIP and Allegro MLIP models.
We find that these optimizers can substantially outperform Adam in both convergence speed and final accuracy. SOAP and SOAP-Muon emerge as robust and consistently strong methods, while Muon only provides partial gains relative to Adam.
The improvements are particularly pronounced under partial force supervision. Our results indicate that optimizer choice is an overlooked yet impactful design axis for MLIPs.
\end{abstract}

\section{Introduction}
Machine learning interatomic potentials (MLIPs)~\cite{Blank1995Neural, Behler2007Generalized,Bartok2010Gaussian} have become an indispensable tool for atomistic simulation in chemistry and materials science~\cite{husistein2026newparadigmcomputationalchemistry}.
They have enabled studies ranging from the thermodynamics of water~\cite{Cheng2019Ab} to the properties of amorphous silicon~\cite{Deringer2021Origins}, large-scale materials screening~\cite{Merchant2023Scaling}, and even all-atom simulations of complex biomolecular assemblies such as the HIV capsid~\cite{Kozinsky2023Scaling}.

Motivated by these successes, the MLIP community has invested heavily in two primary directions of improvement.
The first is \textit{new architectures}, progressing
from descriptor-based~\cite{Bartok2010Gaussian,bartok2013representing,Shapeev_2016, Han_2018,ACE_2019} to deep learning approaches such as invariant~\cite{SCHnet} and equivariant graph neural networks~\cite{Batzner2022E,MACE_2022,Musaelian2023Learning,bochkarev2024graph} and attention-based architectures~\cite{liao2023equiformer,pozdnyakov2023smooth,qu2024importance,qu2026recipe}.
The second direction is \textit{new datasets}, with large-scale quantum chemical databases for organic molecules~\cite{Smith2017ANI, Eastman2023SPICE, Levine2025Open},
catalytic systems~\cite{Chanussot2021Open}, and inorganic
materials~\cite{deng_2023_chgnet, Barroso2024Open} providing the corpora on which these architectures are trained and evaluated.

A third axis of improvement, the \textit{training techniques} that dictate the dynamics of learning, has received comparatively little attention.
Crucially, the choice and configuration of the optimizer used is largely static.
The vast majority of MLIP training pipelines default to Adam~\cite{kingma2017adammethodstochasticoptimization} or one of its closely related variants such as AdamW~\cite{loshchilov2019decoupledweightdecayregularization}.
Even at the scale of training universal potentials, such as the Universal Models for Atoms (UMA)~\cite{wood2025family} and SevenNet-Omni~\cite{kim2026optimizing}, Adam-family optimizers remain the default choice in reported training protocols.

Meanwhile, the scaling of large language models (LLMs) has renewed interest in optimizers that go beyond the diagonal gradient scaling of Adam by exploiting matrix structure in neural-network weight tensors. This broad direction includes explicit structured preconditioners such as Shampoo~\cite{pmlr-v80-gupta18a}, SOAP~\cite{vyas2025soapimprovingstabilizingshampoo}, Kron~\cite{castanyer2026kron}, SPlus~\cite{frans2026a_splus}, and DyKAF~\cite{yudin2025dykafdynamicalkroneckerapproximation}; orthogonalized matrix-update methods such as Muon~\cite{jordan2024muon}, Scion~\cite{pethick2025scion}, Gluon~\cite{riabinin2026gluon}, and PolarGrad~\cite{lau2025polargrad}; and hybrid adaptive schemes such as SOAP-Muon~\cite{vyas2025improving}, COSMOS~\cite{liu2026cosmos}, Mousse~\cite{zhang2026mousserectifyinggeometrymuon}, and Newton-Muon~\cite{du2026newtonmuonoptimizer}.

In this work, we focus on \textbf{Muon} and \textbf{SOAP} as prominent optimizers, together with \textbf{SOAP-Muon} as a hybrid variant combining elements of both.
For LLMs, these methods have demonstrated faster convergence to competitive loss values~\cite{jordan2024muon, Liu2025Muon,vyas2025soapimprovingstabilizingshampoo}. 
However, their potential for MLIP training remains underexplored. \citet{Liu2026Beyond} recently argue that optimizer choice is an overlooked factor in MLIP fine-tuning, but restrict their benchmark to optimizers based on diagonal preconditioning methods. 
\citet{koker2025trainingfoundationmodelmaterials} provided encouraging evidence for this line of inquiry, showing that Muon outperforms Adam as an ingredient for budget-conscious foundation potential training, though without extensive ablations.

We work towards filling this gap by benchmarking Muon, SOAP, and SOAP-Muon against AdamW on two systems of physical significance: liquid water~\cite{Cheng2019Ab}, modeled with NequIP~\cite{Batzner2022E} and the solid acid electrolyte \ce{CsH2PO4} (cesium dihydrogen phosphate, CDP)~\cite{wang2025revealingprotonslingshotmechanism}, important for electrochemical energy technologies, modeled with Allegro~\cite{Musaelian2023Learning}.
We additionally examine optimizer behavior under varying levels of force supervision, including energy-only training.
Computing force labels in density functional theory (DFT) incurs little additional cost beyond the energy evaluation, thanks to the Hellmann–Feynman theorem~\cite{helgaker2013molecular}.
Higher levels of theory, however, do not share this benefit.
For coupled cluster methods, force computation is costly~\cite{smith2019approaching}.
This limitation is shared by diffusion Monte Carlo, where energy-only training is typically used due to the prohibitive cost of force evaluations~\cite{Huang2022Machine}.
Methods that are better suited for sparse-force regimes are thus desirable for training more accurate MLIPs on reference data beyond DFT.
Furthermore, even when forces are available, some works have trained on only a subset to reduce the training time and memory requirements~\cite{Lopez2023net}, highlighting sparse force supervision as a practically relevant regime.

Our contributions here are as follows:
\begin{itemize}
    \item We integrate Muon, SOAP, and SOAP-Muon into the \texttt{nequip} MLIP framework~\cite{Tan2026-fm} and benchmark them against AdamW across two systems. We find that SOAP and SOAP-Muon consistently improve energy and force accuracy and accelerate convergence, with SOAP showing the most robust behavior across systems and SOAP-Muon achieving the strongest results in selected settings. Muon provides gains over AdamW only in one system.
    \item We examine how these optimizers behave under reduced force supervision: SOAP and SOAP-Muon retain strong accuracy as force supervision is reduced; in particular, SOAP-Muon trained with $50\%$ of force labels matches AdamW trained with $100\%$, suggesting a path to reducing force-label requirements in regimes where force labels are expensive.
    \item We demonstrate that the resulting MLIPs are physically faithful, reproducing corresponding \textit{ab initio} calculations and experimental observables. Notably, in one of the systems studied, SOAP-Muon preserves this fidelity even when trained with only $5\%$ of the force labels, while the corresponding AdamW model becomes unstable at the same level of force supervision.
\end{itemize}

\section{Background}
\paragraph{MLIP Training} MLIPs approximate the potential energy surface (PES) of an atomistic system by fitting a neural network with weights $\theta$ to quantum mechanical reference data, most commonly computed using DFT. The network maps a set of atomic positions and chemical species $\{\mathbf{r}_j,Z_j\}$ to a total energy $E_\theta$, which decomposes into a sum of local atomic contributions $\varepsilon_{i, \theta}$ to ensure size extensivity,
\begin{equation}
E_\theta(\{\mathbf{r}_j, Z_j\}) = \sum_i \varepsilon_{i,\theta}(\{\mathbf{r}_j, Z_j\}_{j \in \mathcal{N}_i}),
\end{equation}
where $\mathcal{N}_i$ denotes the local neighborhood of atom $i$ within a cutoff radius. Forces are obtained as the negative gradient of the predicted energy with respect to atomic positions via automatic differentiation,
\begin{equation}
    \mathbf{F}_{i,\theta} = -\frac{\partial E_{\theta}}{\partial \mathbf{r}_i},
\end{equation}
enforcing energy conservation by construction. Training is performed by minimizing a weighted combination of energy and force errors over a dataset of $N$ reference configurations $\{(\{\mathbf{r}_j,Z_j\}_n, E_n, \mathbf{F}_n)\}_{n=1}^N$, where $\mathbf{F}_n \in \mathbb{R}^{3N_\text{atoms}}$ collects the reference forces on all atoms in configuration $n$,
\begin{align} \label{eq:loss}
    \mathcal{L}(\theta) &= \frac{\lambda_E}{N}\sum_{n=1}^N \left(E_\theta(\{\mathbf{r}_j, Z_j\}_n) - E_n\right)^2 \nonumber\\
    &+ \frac{\lambda_F}{3\sum_{n=1}^{N}N_\text{atoms}^{(n)}}\sum_{n=1}^N \left\|\mathbf{F}_\theta(\{\mathbf{r}_j, Z_j\}_n) - \mathbf{F}_n\right\|^2,
\end{align}
where $\lambda_E$ and $\lambda_F$ control the relative contribution of energy and force supervision.
Including forces in the training loss substantially improves accuracy and stability~\cite{Batzner2022E}.
In the force-sparse setting we study here, the force loss is computed over only a subset of configurations, with the limiting case $\lambda_F = 0$ corresponding to energy-only training (see Appendix~\ref{app:training} for details).
\paragraph{NequIP and Allegro} NequIP~\cite{Batzner2022E} and Allegro~\cite{Musaelian2023Learning} are $E(3)$-equivariant graph neural networks, in which internal features are geometric tensors that transform under rotations, reflections, and translations, yielding energy and force predictions that respect the fundamental 3D Euclidean symmetries of atomistic systems.
Local information is mixed across the network through equivariant tensor products.
NequIP propagates features via message-passing between neighboring atoms, while Allegro builds many-body representations through iterated tensor products of atom-pairwise features within a strictly local cutoff.
Together, these architectures laid the foundations for the most widely used class of equivariant MLIPs, known for high accuracy and strong data efficiency attributed to the inductive bias of equivariance~\cite{Batzner2022E, Musaelian2023Learning}. Both are 
implemented within the \texttt{nequip} software framework~\cite{Tan2026-fm}.


\begin{table*}[!t]
  \caption{Test MAE, reported as mean $\scriptstyle\pm$ standard deviation across 5 seeds. Results are shown for the joint energy and force prediction task (E+F) and the energy-only task (E). Gray force entries denote quantities not included in the corresponding training objective. Within each metric and dataset, the best and second-best mean values are indicated by \textbf{boldface} and \underline{underlining}, respectively.}
  \label{tab:main_test_metrics}
  \begin{center}
    \begin{small}
      \begin{sc}
        \begin{tabular}{llcccc}
          \toprule
          \multirow{2}{*}{Task} & \multirow{2}{*}{Optimizer}
            & \multicolumn{2}{c}{CDP (Allegro)} & \multicolumn{2}{c}{Water (NequIP)} \\
          \cmidrule(lr){3-4} \cmidrule(lr){5-6}
            & & E [$\mathrm{meV/atom}$] ($\downarrow$) & F [$\mathrm{meV/\AA}$] ($\downarrow$) & E [$\mathrm{meV/atom}$] ($\downarrow$) & F [$\mathrm{meV/\AA}$] ($\downarrow$) \\
          \midrule
          \multirow{4}{*}{E+F}
            & AdamW     & $0.628 \scriptstyle\pm 0.0434$                    & $32.2 \scriptstyle\pm 0.615$               & $0.773 \scriptstyle\pm 0.0713$                    & $25.7 \scriptstyle\pm 1.44$ \\
            & Muon      & \underline{$0.581 \scriptstyle\pm 0.0328$}        & $29.6 \scriptstyle\pm 0.617$               & $1.53 \scriptstyle\pm 0.317$                      & $26.6 \scriptstyle\pm 2.90$ \\
            & SOAP      & $\mathbf{0.569 \scriptstyle\pm 0.0694}$           & \underline{$29.6 \scriptstyle\pm 0.305$}   & \underline{$0.604 \scriptstyle\pm 0.0105$}        & $\mathbf{20.9 \scriptstyle\pm 0.698}$ \\
            & SOAP-Muon & $0.582 \scriptstyle\pm 0.0469$                    & $\mathbf{27.8 \scriptstyle\pm 0.634}$      & $\mathbf{0.590 \scriptstyle\pm 0.0687}$           & \underline{$21.0 \scriptstyle\pm 1.04$} \\
          \midrule
          \multirow{4}{*}{E}
            & AdamW     & $5.16 \scriptstyle\pm 0.178$                      & \g{$503 \scriptstyle\pm 31.8$}             & $3.21 \scriptstyle\pm 0.272$                      & \g{$306 \scriptstyle\pm 46.1$} \\
            & Muon      & $3.30 \scriptstyle\pm 0.135$                      & \g{$281 \scriptstyle\pm 9.73$}             & $5.37 \scriptstyle\pm 0.813$                      & \g{$591 \scriptstyle\pm 59.1$} \\
            & SOAP      & \underline{$3.03 \scriptstyle\pm 0.389$}          & \g{\underline{$214 \scriptstyle\pm 25.7$}} & $\mathbf{2.38 \scriptstyle\pm 0.573}$             & \g{$\mathbf{236 \scriptstyle\pm 54.8}$} \\
            & SOAP-Muon & $\mathbf{2.75 \scriptstyle\pm 0.163}$             & \g{$\mathbf{201 \scriptstyle\pm 18.8}$}    & \underline{$2.77 \scriptstyle\pm 0.346$}          & \g{\underline{$269 \scriptstyle\pm 24.5$}} \\
          \bottomrule
        \end{tabular}
      \end{sc}
    \end{small}
  \end{center}
\end{table*}

\paragraph{Optimizers} Vanilla gradient descent updates parameters in the direction of steepest descent. However, the loss surface $\mathcal{L}(\theta)$ is generally anisotropic, i.e., some directions change the loss more than others per unit step.
Preconditioning addresses this inconsistency by rescaling the gradient to account for the local geometry of the loss, allowing larger steps in flat directions and smaller steps in steep ones. For a vector-valued parameter $w$ with gradient $g$, a preconditioned update takes the form
\begin{equation}
    w \leftarrow w - \eta P^{-1} g,
\end{equation}
where $P$ is a positive definite matrix describing the local curvature and $\eta$ is the learning rate. A natural choice is the Hessian, but forming, storing, and inverting the Hessian is prohibitively expensive at scale. Practical optimizers therefore choose tractable approximations for $P$.
Additionally, most optimizers incorporate momentum, an exponential moving average of past gradients, to smooth noisy updates and accelerate convergence.

\textbf{Adam}~\cite{kingma2017adammethodstochasticoptimization} approximates $P$ with a diagonal matrix $P = \operatorname{diag} (\hat{v}_t)^{1/2}$, where $\hat{v}_t$ is a running estimate of the element-wise squared gradient. \textbf{AdamW}~\cite{loshchilov2019decoupledweightdecayregularization} decouples the weight decay from the Adam gradient update. 

For a matrix-valued parameter $W$ with gradient $G \in \mathbb{R}^{m \times n}$, the full
preconditioner on $\text{vec}(W)$ would be an $mn \times mn$ matrix, which is intractable in practice. An approximation is to assume Kronecker structure $P \approx L^\mathsf{T}\otimes R$, which leads to the update
\begin{equation}
    W \leftarrow W - \eta L^{-p}GR^{-p},
\end{equation}
where $L \in \mathbb{R}^{m \times m}$ and $R \in \mathbb{R}^{n\times n}$ are left and right preconditioners, and $p$ is the power of the preconditioner.

\textbf{Shampoo}~\cite{pmlr-v80-gupta18a} defines the left and right preconditioners using the accumulated uncentered row and column covariances of the gradient matrix, updating them as $L \leftarrow L + GG^\mathsf{T}$ and $R \leftarrow R + G^\mathsf{T}G$. The preconditioning power is set to $p=1/4$. Later work by \citet{bernstein2024oldoptimizernewnorm} showed that, when the accumulation is removed, Shampoo simplifies to a semi-orthogonal weight update. In this regime, the method can be interpreted as projecting the gradient matrix onto the nearest semi-orthogonal matrix under the Frobenius norm.

\textbf{Muon}~\cite{jordan2024muon} builds on this perspective, by first applying a Nesterov-style momentum update and then performing the orthogonalization step using efficient Newton--Schulz iterations.

\textbf{SOAP}~\cite{vyas2025soapimprovingstabilizingshampoo} is motivated by the observation that Shampoo is equivalent to running Adafactor~\cite{DBLP:journals/corr/abs-1804-04235} in the eigenspace of the preconditioner. That is, the gradient is transformed as $G_t'\leftarrow Q^\mathsf{T}_L G Q_R$ where $Q_L$ and $Q_R$ are the eigenvector matrices of the $L$ and $R$, respectively. SOAP replaces the Adafactor update in this eigenspace with an AdamW step, and then projects the result back to the original parameter space.

\textbf{SOAP-Muon}~\cite{vyas2025improving} extends SOAP by adding a Muon-inspired orthogonalization step after the standard SOAP step. 




\begin{figure*}[!t]
\centering
\includegraphics[width=0.93\linewidth]{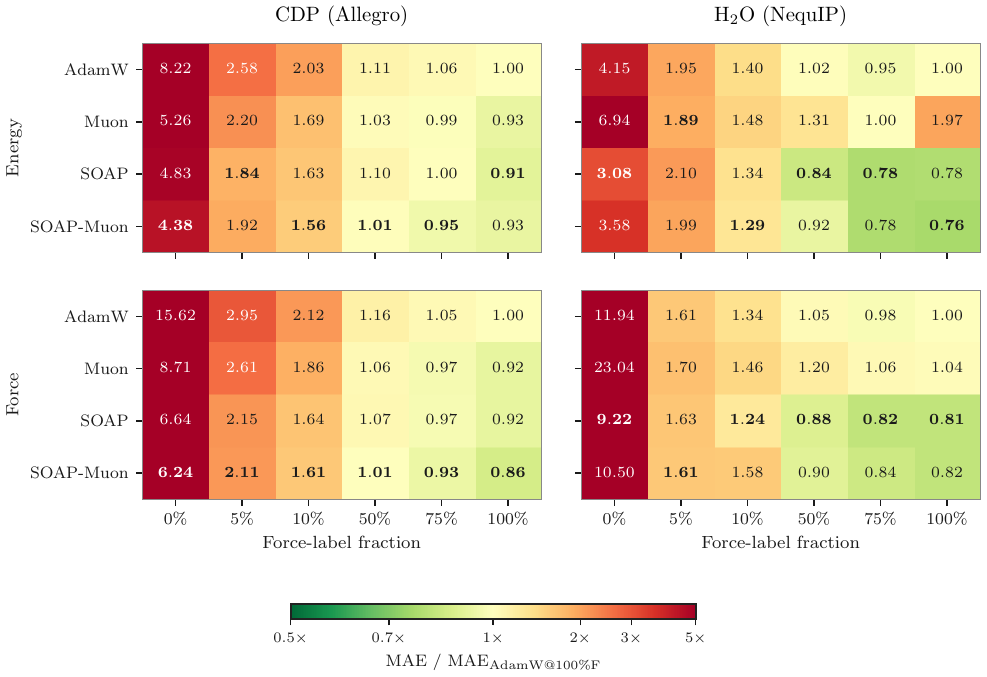}
\caption{Relative test MAE vs AdamW at full ($100\%$) force supervision. Each cell corresponds to $\text{MAE}_{\text{optimizer}} / \text{MAE}_{\text{AdamW@100\%F}}$. Green cells are below AdamW@100\%F (better), red cells above. \textbf{Boldface} values indicate the best (lowest) MAE within each force-\% column.}
\label{fig:main-heatmap}
\end{figure*}

\begin{figure*}[tb]
\centering
\includegraphics[width=0.95\linewidth]{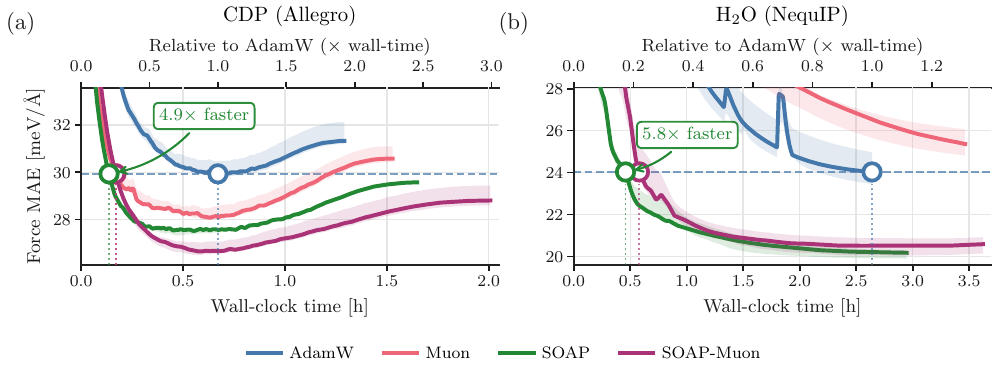}
\caption{Wall-clock convergence under full force supervision. Validation force MAE versus wall-clock time on CDP (a) and water (b). Lines are per-epoch medians and shaded bands are interquartile ranges across three seeds, all executed on NVIDIA A100 GPUs. The dashed line marks AdamW’s minimum median force MAE and circles indicate the first wall-clock time at which each optimizer's median curve crosses that level.}
\label{fig:main-walltime}
\end{figure*}

\section{Experimental Setup}
To assess the versatility of the studied optimizers, we evaluate them across different equivariant architectures and chemical environments. We consider two representative settings: homogeneous liquid water~\cite{Cheng2019Ab} modeled with NequIP, and solid-state CDP~\cite{wang2025revealingprotonslingshotmechanism} modeled with Allegro. 

Our chosen model configurations are based on previously reported parameters~\cite{Batzner2022E,wang2025revealingprotonslingshotmechanism}. To ensure a fair comparison across each model, dataset, and optimizer, we use a systematic hyperparameter tuning protocol (see Appendix~\ref{app:training} for details).

To probe the limits of each optimizer, we additionally consider reduced force supervision, including the energy-only regime. This setting evaluates how these optimizers perform with reduced data. Accordingly, we repeat the tuning protocol across training sets with varying fractions of force-labeled frames---$75\%$, $50\%$, $10\%$, $5\%$, and $0\%$ (energy-only training)---while retaining all energy labels.

We evaluate the resulting models using per-atom energy and force mean absolute error (MAE), as well as training wall-clock time. However, energy and force error metrics alone do not fully capture the quality of an MLIP~\cite{fu2022forces}. We therefore further assess the physical fidelity of the optimal PES identified by each optimizer by performing molecular dynamics (MD) simulations using the trained MLIPs and comparing resulting observables against experimental measurements and \textit{ab initio} reference data.

\section{Results}
\subsection{Full Supervision}

\paragraph{Accuracy Gains}
Under full energy and force supervision, the matrix-structured optimizers studied substantially outperform AdamW, with the magnitude and consistency of the gains depending on the system (Table~\ref{tab:main_test_metrics}, Figure~\ref{fig:main-heatmap}). On CDP with Allegro, all three optimizers improve over AdamW. SOAP achieves the best energy accuracy, reducing the energy MAE by $9\%$
whereas SOAP-Muon achieves the best force accuracy, reducing the MAE by $14\%$.
On water with NequIP, SOAP and SOAP-Muon again yield lower MAEs than AdamW. In this case, however, SOAP-Muon gives the best energy accuracy, reducing the MAE by $24\%$
while SOAP provides the best force accuracy, reducing the MAE by $19\%$.
Muon alone underperforms the AdamW baseline, particularly for energy prediction.

\paragraph{Speedup} The accuracy improvements are accompanied by substantial reductions in time-to-accuracy. We compare the wall-clock time required for each optimizer to reach AdamW’s minimum median validation force MAE under full force supervision (Figure~\ref{fig:main-walltime}). On CDP, SOAP reaches that target $4.9\times$ faster than AdamW, while on water it does so $5.8\times$ faster. These gains are notable because matrix-structured methods incur additional computation per optimization step relative to AdamW. The wall-clock advantage therefore implies that they reduce the number of epochs needed by an even larger margin. In practice, improved conditioning more than compensates for the higher per-step cost, making SOAP attractive not only for final accuracy but also for training efficiency.

\subsection{Reduced Supervision}
\paragraph{Energy-Only Supervision} In the energy-only training regime, where forces are excluded from the training objective and therefore provide a stringent test of whether the learned PES captures physically meaningful gradients, the advantage of matrix-structured optimizers becomes more pronounced.
On CDP, all three alternatives substantially outperform AdamW.
SOAP-Muon achieves the strongest overall results, reducing the energy and force MAE by roughly $47\%$ and $60\%$.
On water, SOAP performs best, reducing both energy and force errors relative to AdamW by $26\%$ and $23\%$, while Muon performs noticeably worse.

\paragraph{Sparse Force Supervision} The same ranking largely persists as the number of force-labeled configurations decreases, with the gains over AdamW becoming especially valuable in the low-label regime (Figure~\ref{fig:main-heatmap}). On CDP, SOAP-Muon trained with only $50\%$ force supervision achieves accuracy comparable to AdamW trained with the full force-labeled set, and SOAP shows a similar pattern. Both methods also remain clearly stronger than AdamW at $10\%$ and $5\%$ force supervision. This suggests that improved optimization can partially compensate for reduced access to force labels, lowering the amount of expensive force data needed to reach a given level of accuracy. On water, the trend is less regular, with some models trained at $75\%$ and $50\%$ force supervision often outperforming their respective models trained on $100\%$. Nevertheless, SOAP is the most consistently competitive method across supervision levels and is the strongest single default in our experiments.

\subsection{Physical Fidelity}
To assess whether the optimizer-dependent differences in MAE translate into meaningful differences in the learned PESs, we evaluated the trained MLIPs in MD simulations. 
\paragraph{CDP} We followed the MD protocol of \citet{wang2025revealingprotonslingshotmechanism}. All fully force-supervised models faithfully reproduce the \textit{ab initio} MD (AIMD) results, matching the radial distribution functions (RDFs) and showing consistent mean squared displacement (MSD) behaviors across optimizers (Appendix Figure~\ref{fig:CDPfid}). The differences become much more pronounced in the sparse-label regime. Notably, at $5\%$ force supervision, SOAP-Muon remains stable and accurately reproduces the AIMD RDFs and MSD. This is in sharp contrast to AdamW that exhibits catastrophic instability, with trajectories diverging almost immediately and producing nonphysical results under the same conditions, as shown in Figure~\ref{fig:main_CDP} (the full set of RDFs and MSD is reported in Appendix Figure~\ref{fig:CDPsparse}). Moreover, the $5\%$ SOAP-Muon model reproduces experimental results, yielding a proton diffusion activation energy of $E_a = 0.42\;\mathrm{eV}$ (Appendix Figure~\ref{fig:CDPsparse}h), in good agreement with the experimental range of $E_a = 0.39\text{--}0.43\;\mathrm{eV}$ \cite{Haile2007Solid,Ishikawa2008Proton,wang2025gradientalignmentphysicsinformedneural}.

\begin{figure}[tb]
  \vskip 0.2in
  \begin{center}
    \centerline{\includegraphics[width=\columnwidth]{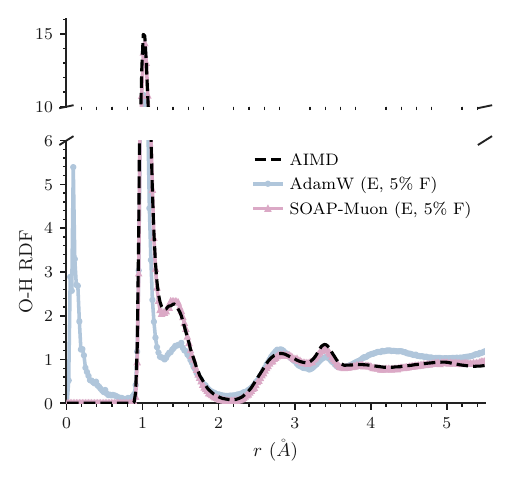}}
    \caption{Radial distribution functions (RDFs) of O–H pairs in CDP obtained from MD simulations using AdamW (blue) and SOAP-Muon (pink) models trained on energies and $5\%$ of forces, compared against AIMD ground truth (black).}
    \label{fig:main_CDP}
  \end{center}
\end{figure}

\paragraph{Water} The fully force-supervised models also generate similar structural and dynamical observables across optimizers, with RDFs and MSDs broadly consistent with one another and with the experimental O-O RDF reference (Appendix Figure~\ref{fig:H2Ofid})~\cite{Skinner2014Structure}. For all sparsity percentages tested, both AdamW and SOAP maintained physical fidelity on RDFs and MSD (Appendix Figure~\ref{fig:H2Osparse}).

\section{Discussion}
The two cases we study differ both in model architecture (NequIP vs.\ Allegro) and in chemical character (liquid water vs.\ multicomponent solid CDP). 
On CDP, all three matrix-based optimizers outperform AdamW, with a clear progression from AdamW to Muon, SOAP and SOAP-Muon. 
On water, SOAP and SOAP-Muon also improve upon AdamW, with SOAP having a slight advantage; by contrast, Muon performs consistently worse than AdamW across all force-label fractions tested. We note, however, that unlike CDP, the water case required additional tuning of SOAP-Muon (momentum coefficients and singular-value power; see Appendix~\ref{app:training} for details) to achieve these results.
Taken together, SOAP and SOAP-Muon outperform AdamW across most metrics, training settings, and systems, with SOAP showing the most robust behavior across systems. For practitioners choosing a single default optimizer to use instead of AdamW without further tuning, our results suggest SOAP. 

Muon's underperformance on water is a notable result that warrants further investigation. The behavior across all three matrix-structured variants suggests that the orthogonalization
step is the primary source of degradation, with adaptive preconditioning mitigating but not eliminating its effects. Similar concerns about the stability of Muon have been raised in the context of physics-informed neural networks~\cite{Lu2026Muon}, where step-size
regulation along dominant spectral modes was proposed as a potential remedy.

\paragraph{Limitations}
Our work is restricted to training on two systems and two equivariant architectures. Foundation potential training and fine-tuning are interesting open directions, and we expect SOAP to remain effective at the present scale of foundation models, though this requires a systematic study.

\section{Conclusion}
Across the systems and training settings studied, matrix-structured optimizers consistently outperform AdamW for MLIP training, with SOAP emerging as the most robust default with the advantage compounding under sparse force supervision.
These accuracy gains translate into substantial wall-clock speedups despite per-step preconditioner overhead, and into faithful molecular-dynamics behavior including observables in good agreement with experiment.
We argue that training techniques, exemplified by optimizer choice in this work, should be treated as a first-class design axis in MLIP training, alongside architecture and dataset, particularly as the field moves toward universal foundation potentials where label efficiency and convergence speed become increasingly important.

\section*{Code Availability}
All optimizer code used in this work will be released in future versions of the open-source \texttt{nequip} framework and associated repositories, including \texttt{github.com/mir-group/nequip} and \texttt{github.com/mir-group/allegro}.

\section*{Acknowledgments}
The authors gratefully acknowledge Menghang Wang for valuable insights and support regarding the CDP calculations.  
The computations in this paper were run on the FASRC Cannon cluster supported by the FAS Division of Science Research Computing Group at Harvard University.
An award of computer time was provided by the INCITE program. This research used resources of the Oak Ridge Leadership Computing Facility, which is a DOE Office of Science User Facility supported under Contract DE-AC05-00OR22725.
L.Z. was supported by the National Science Foundation Graduate Research Fellowship under Grant No. DGE-2140743.
This work was supported by the National Science Foundation through the Harvard University Materials Research Science and Engineering Center Grant No. DMR-2011754.

\section*{Impact Statement}

This paper presents work whose goal is to advance the field of Machine
Learning. There are many potential societal consequences of our work, none
which we feel must be specifically highlighted here.

\bibliography{example_paper}
\bibliographystyle{icml2026}

\newpage
\appendix
\onecolumn
\section{Optimizer Details}
\label{app:optimizer_details}

\subsection{AdamW}
We use the PyTorch \texttt{torch.optim.AdamW} implementation.

\subsection{Muon} \label{subsec:muon}
Algorithm~\ref{alg:eq_muon} summarizes the Muon update for a matrix-valued parameter. The key operation is the orthogonalization of a momentum-augmented gradient matrix. In practice, this orthogonalization is approximated using the Newton--Schulz algorithm (denoted $\mathrm{NewtonSchulz5}(\cdot)$ in Algorithm~\ref{alg:eq_muon}), an iterative matrix procedure for approximating the orthogonal factor $UV^\top$ in the decomposition $G = U\Sigma V^\mathsf{T}$, i.e., the matrix obtained by replacing the singular values of $G$ with ones. Because this orthogonalization is defined for two-dimensional weight tensors (and 4-dimensional convolutional parameters), non-matrix parameters are assigned to an auxiliary AdamW optimizer (for simplicity will be referred to as Adam). ~\citet{jordan2024muon} further observed empirically that embedding and readout weights are also better optimized with Adam.

Our \texttt{nequip} implementation builds on the official Muon reference implementation and retains the original $\beta$ momentum hyperparameter. To accommodate the split between parameters partitioned to Muon and those handled by the auxiliary Adam optimizer, we define two parameter groups, one for each optimizer. Broadly following the recommendations of Jordan et al.~\cite{jordan2024muon}, we assign intermediate-layer linear weight matrices to Muon, while optimizing the type embeddings and readout layers with Adam. A complete breakdown of the parameter groups is provided in Table~\ref{tab:nequip-allegro-parameter-groups}.


To implement this split in NequIP, we must also account for the internal parameterization of \texttt{e3nn} layers.
In particular, \texttt{e3nn} stores the weights of its equivariant operators as flattened one-dimensional parameter vectors, even when those weights act as structured matrices or tensors in the forward pass.
Similar to the Nequix implementation~\cite{koker2025trainingfoundationmodelmaterials}, for the subset of \texttt{e3nn}'s \texttt{Linear} parameters assigned to Muon, we first recover the corresponding structured blocks from the flattened representation, perform the Muon update on those reshaped blocks, and then flatten and reassemble them into the original parameter vector. We formalize the instruction-wise action of the \texttt{e3nn} operators below.

\begin{table*}[!b]
\caption{Parameter classes, shapes, and optimizer-group assignments for NequIP and Allegro. \texttt{e3nn.o3.Linear} (\texttt{Linear}) and \texttt{e3nn.o3.FullyConnectedTensorProduct} (\texttt{FCTP}) layers store weights as flattened 1D parameters, which are partitioned and reshaped into structured blocks during use.}
\label{tab:nequip-allegro-parameter-groups}
\centering
\begin{threeparttable}
\small
\setlength{\tabcolsep}{5pt}
\renewcommand{\arraystretch}{1.1}
\begin{sc}
\begin{tabular}{cllc}
\toprule
\multicolumn{1}{c}{Model} &
\multicolumn{1}{c}{Parameter class} &
\multicolumn{1}{c}{Shape} &
\multicolumn{1}{c}{Parameter} \\
&
&
&
\multicolumn{1}{c}{group} \\
\midrule

\multirow{6}{*}{NequIP}
& Type embedding
& \codecell{\texttt{[num\_species, num\_features]}}
& Adam \\
& Edge-MLP weight matrices
& 2D weight matrices
& Muon \\
& \codecell{\texttt{Linear}} weights
& \codecell{\texttt{[weight\_numel]}} ($(m_j,n_j)$ blocks)
& Muon \\
& \codecell{\texttt{FCTP}} weights
& \codecell{\texttt{[weight\_numel]}} ($(m^{(1)}_j,m^{(2)}_j,n_j)$ blocks)
& Adam \tnote{1} \\
& Energy readout MLP
& \codecell{\texttt{[num\_features, 1]}}
& Adam \\
& Per-type energy scale and shift
& \codecell{\texttt{[num\_species, 1]}}
& Adam \\

\midrule

\multirow{6}{*}{Allegro}
& Type embeddings
& \codecell{\texttt{[num\_species, num\_tensor\_features]}}
& Adam \\
& Scalar / tensor embedding MLP weights
& 2D weight matrices
& Muon \\
& Latent MLP weights
& 2D weight matrices
& Muon \\
& First-layer environment MLP weights
& 2D weight matrices
& Muon \\
& Tensor-product weights
& \codecell{\texttt{[num\_tensor\_features, num\_paths]}}
& Adam \\
& Readout MLP weights
& \codecell{\texttt{[num\_scalar\_features, 1]}}
& Adam \\

\bottomrule
\end{tabular}
\end{sc}

\begin{tablenotes}[flushleft]
\footnotesize
\item[1] In future work, these weights could be represented as 2D matrices by flattening the two input dimensions (i.e., $m^{(1)}_j \times m^{(2)}_j \times n^{\mathrm{out}}_j \mapsto
\bigl(m^{(1)}_j m^{(2)}_j\bigr) \times n^{\mathrm{out}}_j)$, analogous to Muon’s treatment of convolutional weights. 
\end{tablenotes}

\end{threeparttable}
\end{table*}

\paragraph{\texttt{e3nn} and tensor-product weights.}
We consider two \texttt{e3nn.o3} layers: (1) \texttt{Linear} and (2) \texttt{FullyConnectedTensorProduct} (\texttt{FCTP}).

In \texttt{Linear}, each instruction connects an input irrep block to an output irrep block of the same irrep type. The learnable weight for instruction $j$ is a multiplicity-mixing matrix
\begin{equation}
    W_j \in \mathbb{R}^{m^{\mathrm{in}}_j \times n^{\mathrm{out}}_j},
\end{equation}
which mixes copies of the irrep but does not act on the internal $2\ell+1$-dimensional irrep basis.

In \texttt{FCTP}, each instruction combines an irrep block from the first input, an irrep block from the second input, and an admissible output irrep block. The learnable weight for instruction \(j\) is a multiplicity-mixing tensor
\begin{equation}
    T_j \in \mathbb{R}^{m^{(1)}_j \times m^{(2)}_j \times n^{\mathrm{out}}_j},
\end{equation}
which mixes copies of the two input irreps and the output irrep but does not act directly on the internal $2\ell+1$-dimensional irrep bases; those couplings are fixed by the tensor-product Clebsch--Gordan structure.


Thus, \texttt{Linear} uses learned 2D multiplicity-mixing blocks, while \texttt{FCTP} uses learned 3D blocks over the two input multiplicities and the output multiplicity.

\paragraph{Flattened storage.}
Although these operators act through structured blocks $W_j$ or $T_j$, \texttt{e3nn} stores all learnable weights as a single flattened parameter vector
\begin{equation}
    w \in \mathbb{R}^{P}, \qquad P = \sum_{j=1}^{J} |\mathbf{s}_j|,
\end{equation}
where $|\mathbf{s}_j|$ is the number of entries in instruction $j$'s block. Equivalently, we define a slice specification
\begin{equation}
    \mathcal{S} = \{(\mathcal{I}_j,\mathbf{s}_j)\}_{j=1}^{J},
\end{equation}
with $\mathcal{I}_j$ selecting the entries of $w$ belonging to instruction $j$, and
\begin{equation}
    B_j = \operatorname{reshape}(w_{\mathcal{I}_j}, \mathbf{s}_j),
\end{equation}
where $B_j = W_j$ and $\mathbf{s}_j = (m^{\mathrm{in}}_j,n^{\mathrm{out}}_j)$ for \texttt{Linear}, while $B_j = T_j$ and $\mathbf{s}_j = (m^{(1)}_j,m^{(2)}_j,n^{\mathrm{out}}_j)$ for \texttt{FCTP}. The forward pass uses the reshaped blocks $B_j$. Adam-style elementwise optimizers can operate directly on the flattened parameter vector $w$, because their updates do not depend on the structural interpretation of the parameter shape. By contrast, matrix-structured optimizers such as Muon, SOAP, and SOAP-Muon are shape-sensitive. To use their structured orthogonalization or preconditioning, the relevant slices of $w$ must be reshaped into their instruction-wise matrix or tensor blocks, updated in that form, and then flattened back into the original parameter vector.

\begin{algorithm}[t]
\caption{Muon optimizer update~\cite{jordan2024muon}}
\label{alg:eq_muon}
\begin{algorithmic}[1]
\REQUIRE Parameter matrix $W \in \mathbb{R}^{m \times n}$, step size $\eta$, $\beta$
\STATE $M_0 \gets 0$
\FOR{$t = 1, 2, \ldots$}
    \STATE $G_t \gets \nabla_W \mathcal{L}_t(W_{t-1}) \in \mathbb{R}^{m \times n}$ \COMMENT{Current gradient}
    \STATE $M_t \gets \beta M_{t-1} + (1-\beta) G_t$ \COMMENT{Momentum buffer}
    \STATE $U_t \gets \beta M_t + (1-\beta) G_t$ \COMMENT{Nesterov-style momentum}
    \STATE $O_t \gets \mathrm{NewtonSchulz5}(U_t)$ \COMMENT{Orthogonalized direction}
    \STATE $W_t \gets W_{t-1} - \eta O_t$ \COMMENT{Parameter update}
\ENDFOR
\RETURN $W_t$
\end{algorithmic}
\end{algorithm}

\subsection{SOAP}
Algorithm~\ref{alg:eq_soap_muon_optional} presents the general SOAP update step, simplified by omitting the initialization steps and AdamW weight decay. For the standard SOAP setup, the orthogonalization flag $\texttt{ortho}$ and normalization flag $\texttt{normalize}$ are turned off. We use the default hyperparameters, $\beta_1=0.95$, $\beta_2=0.95$, Shampoo $\beta=\beta_2=0.95$ and preconditioning frequency of $10$~\cite{vyas2025soapimprovingstabilizingshampoo}.

Unlike AdamW, whose elementwise updates are largely insensitive to how a parameter tensor is reshaped, SOAP uses Shampoo-style structured preconditioning along tensor modes. Consequently, the tensor shape assigned to each parameter block affects the preconditioner and the resulting update.

To leverage this preconditioning, the \texttt{e3nn} linear weights are unstacked and restored to their original two-dimensional forms, following the methodology described in Section~\ref{subsec:muon}. In the present study, the \texttt{e3nn} fully connected tensor product weights were not sliced and reshaped into three-dimensional tensors. We leave the implementation of such structural modifications for future investigations.

\begin{algorithm}[t]
\caption{SOAP update with optional Muon-style orthogonalization and normalization~\cite{vyas2025soapimprovingstabilizingshampoo,vyas2025improving}.}
\label{alg:eq_soap_muon_optional}
\begin{algorithmic}[1]
\REQUIRE Matrix block $W \in \mathbb{R}^{m \times n}$, learning rate $\eta$, betas $(\beta,\beta_1,\beta_2)$, $\epsilon$, preconditioning frequency $f$, orthogonalization flag $\texttt{ortho}$, normalization flag $\texttt{normalize}$,  singular-value power $\rho$
\FOR{$t = 1,2,\ldots$}
    \STATE $G_t \leftarrow \nabla_W \mathcal{L}_t(W_{t-1}) \in \mathbb{R}^{m \times n}$ \COMMENT{Current gradient}
    \STATE $\widetilde{G}_t \leftarrow Q_{L,t-1}^\mathsf{T} G_t Q_{R,t-1}$ \COMMENT{Gradient projection to Shampoo eigenbasis}
    \STATE $M_t \leftarrow \beta_1 M_{t-1} + (1-\beta_1)\widetilde{G}_t$ \COMMENT{(Adam) First moment estimate}
    \STATE $V_t \leftarrow \beta_2 V_{t-1} + (1-\beta_2)(\widetilde{G}_t \odot \widetilde{G}_t)$ \COMMENT{(Adam) Second moment estimate}
    \STATE $\alpha_t \leftarrow \eta \sqrt{1-\beta_2^t}/(1-\beta_1^t)$ \COMMENT{(Adam) Bias-corrected step size}
    \STATE $\widetilde{U}_t \leftarrow M_t \oslash (\sqrt{V_t} + \epsilon)$ \COMMENT{(Adam) Elementwise variance-normalization}
    \STATE $U_t \leftarrow Q_{L,t-1}\widetilde{U}_t Q_{R,t-1}^\mathsf{T}$ \COMMENT{Back projection}

    \IF{$\texttt{ortho}$}
        \IF{$\rho = 0$}
            \STATE $U_t \leftarrow \mathrm{NewtonSchulz5}(U_t)$ \COMMENT{(Muon) orthogonalization via Newton--Schulz iteration}
        \ELSIF{$\rho \neq 1$}
            \STATE Compute SVD $U_t = P\Sigma R^\mathsf{T}$ \COMMENT{(Muon) spectral decomposition}
            \STATE $U_t \leftarrow P\Sigma^\rho R^\mathsf{T}$ \COMMENT{(Muon) singular-value power transform}
        \ENDIF
    \ENDIF

    \IF{$\texttt{normalize}$}
        \STATE $U_t \leftarrow U_t / \sqrt{\mathrm{mean}(U_t^2)}$ \COMMENT{(Muon) RMS normalization}
    \ENDIF

    \STATE $L_t \leftarrow \beta L_{t-1} + (1-\beta) G_t G_t^\mathsf{T}$ \COMMENT{Left preconditioner update}
    \STATE $R_t \leftarrow \beta R_{t-1} + (1-\beta) G_t^\mathsf{T} G_t$ \COMMENT{Right preconditioner update}
    \IF{$t \,\%\, f = 0$} 
        \STATE $S_L \leftarrow L_t Q_{L,t-1}$; $S_R \leftarrow R_t Q_{R,t-1}$ \COMMENT{Power iteration}
        \STATE $Q_{L,t} \leftarrow \texttt{QR}(S_L)$; $Q_{R,t} \leftarrow \texttt{QR}(S_R)$ \COMMENT{Update Shampoo eigenbasis}
    \ELSE 
        \STATE $Q_{L,t} \leftarrow Q_{L,t-1}$; $Q_{R,t} \leftarrow Q_{R,t-1}$ \COMMENT{Keep previous Shampoo eigenbasis}
    \ENDIF

    \STATE $W_t \leftarrow W_{t-1} - \alpha_t U_t$ \COMMENT{Parameter update}
\ENDFOR
\RETURN $W_t$
\end{algorithmic}
\end{algorithm}

\subsection{SOAP-Muon}
Algorithm~\ref{alg:eq_soap_muon_optional} presents the general SOAP update step with the orthogonalization flag $\texttt{ortho}$ and the normalization flag $\texttt{normalize}$ both enabled for SOAP-Muon (initialization and AdamW-style weight decay omitted for simplicity). The resulting update combines Muon-style orthogonalization with RMS normalization of the back-projected AdamW update direction. We use the default preconditioner update frequency of 10 and apply full preconditioning, in which the preconditioner is applied along every mode of the weight tensor. Throughout, the Shampoo preconditioner $\beta$ was set at $\beta=\beta_2$.

A key hyperparameter in this update is the singular-value power $\rho$, which is applied to the singular-value matrix during the orthogonalization step. The official implementation uses the default value $\rho=0.5$, which requires a full SVD and is therefore computationally more expensive. A cheaper Muon-style alternative is to set $\rho=0$, which permits the use of Newton--Schulz iteration (as the transform reduces to $UV^\mathsf{T}$). However, \citet{vyas2025improving} reported that using $\rho=0$ can lead to dataset-dependent training instability. We therefore treat $\rho=0.5$ as the safer default, while using $\rho=0$ when training remains stable, since this enables the more efficient Newton--Schulz-based orthogonalization.

Finally, we use the parameter grouping scheme described in Appendix~\ref{subsec:muon} to determine which parameters receive Muon-style orthogonalization, while the remaining parameters receive standard SOAP updates without the orthogonalization. This grouping also allows us to assign separate $\beta_1,\beta_2$ hyperparameters to the orthogonalized Muon group and the non-orthogonalized Adam group. This differs from the official SOAP-Muon optimizer, which shares the same momentum settings across all parameters.

\section{Training Setup} \label{app:training}

\paragraph{Sparse Force Training} 
\label{sparse_setup}
Given a desired force sparsity level $s \in [0,1]$, we retain
$N_{\mathrm{keep}} = \operatorname{round}(sN)$ 
frames with force labels, where $N$ denotes the total number of frames in the corresponding dataset split. 
In practice, for each sparsity level, the force-labeled frames are sampled uniformly at random once at training initialization.
We then define a binary mask $m_n \in \{0,1\}$ by setting $m_n = 1$ for the retained frames and $m_n = 0$ for the remaining frames, whose force labels are discarded. 
The same sampled force-labeled subset is used across optimizers for a given data split and random seed.

Under this setting, the loss in Eq.~\ref{eq:loss} becomes
\begin{align}
    \mathcal{L}(\theta) &= \frac{\lambda_E}{N}\sum_{n=1}^N \left(E_\theta(\{\mathbf{r}_j, Z_j\}_n) - E_n\right)^2 + \frac{\lambda_F}{3\sum_{n=1}^{N}m_nN_\text{atoms}^{(n)}}\sum_{n=1}^N m_n\left\|\mathbf{F}_\theta(\{\mathbf{r}_j, Z_j\}_n) - \mathbf{F}_n\right\|^2.
\end{align}
Thus, the force term is normalized only by the number of force components in the force-labeled configurations, rather than by all configurations in the split.
The coefficients $\lambda_E$ and $\lambda_F$ are kept fixed across sparsity levels using the ratio tuned under full force supervision.
The same masking procedure is applied to the training and validation subsets, whereas the test subset is left unmasked. 

\paragraph{Model Architectures.} The model architectures used in this work are based on previously reported configurations for Allegro on CDP and NequIP on water, which were originally optimized using Adam. The architecture hyperparameters are presented in Table~\ref{tab:architecture_hyperparams}.

\begin{table*}[p]
  \centering
  \caption{Model architecture hyperparameters used for the CDP and water experiments.}
  \label{tab:architecture_hyperparams}

  \begin{subtable}[t]{0.45\textwidth}
    \centering
    \caption{CDP experiments with Allegro~\cite{wang2025revealingprotonslingshotmechanism}.}
    \label{tab:cdp_allegro_architecture_hyperparams}
    \begin{footnotesize}
    \begin{sc}
    \begin{tabularx}{\linewidth}{@{}>{\raggedright\arraybackslash}X l@{}}
      \toprule
      Hyperparameter & Value \\
      \midrule
      Cutoff radius & $7.0 \; \mathrm{\AA}$ \\
      Number of layers & $2$ \\
      $l_{\max}$ & $2$ \\
      Parity & True \\
      Number of scalar features & $32$ \\
      Number of tensor features & $32$ \\
      Radial Bessel basis functions & $8$ \\
      Trainable Bessel basis & False \\
      Polynomial cutoff exponent & $6$ \\
      Radial-chemical embedding dimension & $32$ \\
      Scalar embedding MLP depth & $2$ \\
      Scalar embedding MLP width & $128$ \\
      Scalar embedding nonlinearity & SiLU \\
      Allegro MLP depth & $2$ \\
      Allegro MLP width & $128$ \\
      Allegro MLP nonlinearity & SiLU \\
      Tensor-product path-channel coupling & False \\
      Readout MLP depth & $1$ \\
      Readout MLP width & $32$ \\
      Readout MLP nonlinearity & None \\
      Per-type energy scales trainable & True \\
      Per-type energy shifts trainable & True \\
      \bottomrule
    \end{tabularx}
    \end{sc}
    \end{footnotesize}
  \end{subtable}
  \hfill
  \begin{subtable}[t]{0.45\textwidth}
    \centering
    \caption{Water experiments with NequIP~\cite{Batzner2022E}.}
    \label{tab:water_nequip_architecture_hyperparams}
    \begin{footnotesize}
    \begin{sc}
    \begin{tabularx}{\linewidth}{@{}>{\raggedright\arraybackslash}X l@{}}
      \toprule
      Hyperparameter & Value \\
      \midrule
      Cutoff radius & $4.5 \; \mathrm{\AA}$ \\
      Number of interaction layers & $4$ \\
      $l_{\max}$ & $2$ \\
      Parity & True \\
      Number of features & $32$ \\
      Radial Bessel basis functions & $8$ \\
      Trainable Bessel basis & False \\
      Polynomial cutoff exponent & $6$ \\
      Radial MLP depth & $3$ \\
      Radial MLP width & $64$ \\
      Per-type energy scales trainable & True \\
      Per-type energy shifts trainable & True \\
      Pair potential & ZBL \\
      \bottomrule
    \end{tabularx}
    \end{sc}
    \end{footnotesize}
  \end{subtable}
\end{table*}

\paragraph{Hyperparameter Search} 
First, a grid search was performed for the learning rate and weight decay. For each optimizer and task, we first identified a candidate learning-rate order of magnitude $10^{-p}$, and then swept over
$\{3\times10^{-(p+1)}, 10^{-p}, 3\times10^{-p}\}$ while weight decay was sampled across four values $\{0, 10^{-5}, 10^{-4}, 10^{-3}\}$. 
For the water dataset with NequIP, which proved more sensitive to learning rate, we included an additional value of $5\times10^{-p}$.
In Muon and SOAP-Muon, the learning rate and weight decay for both parameter groups were fixed together, although they can be further optimized separately. 
With these parameters fixed, then, for the full force-supervision setting ($100\%$), the energy-force loss coefficient ratio was tuned by sweeping through $\{1:1, 1:10, 1:100\}$. 
Final configurations were selected using a validation metric given by a weighted sum of the per-atom energy and force errors, with a $10$:$1$ weighting between the two components.. 
For NequIP on the water dataset, we used a reduce on plateau scheduler following the settings in Batzner et al.~\cite{Batzner2022E}. 
For Allegro on CDP, we opted for a different approach and used a cosine annealing scheduler. All hyperparameter sweeps were run with a single random seed and shortened training schedules: 100 epochs for CDP and 500 epochs for water.

\paragraph{Training Runs}
The full training runs were conducted using five random seeds (7, 42, 123, 2026, and 1618). The best optimizer hyperparameters identified in the sweeps, and subsequently used in the training runs, are reported in Table~\ref{tab:hyperparameters}. Training was run for 1000 epochs for CDP and 2000 epochs for water. Early stopping was applied using the weighted validation objective. The minimum improvement thresholds were $10^{-6}$ for water and $10^{-5}$ for CDP, with patience values of 600 and 500 epochs, respectively. Training was conducted on NVIDIA A100-SXM4-80GB and H200 GPUs. To ensure consistency, all wall-clock time comparisons were derived exclusively from training runs performed on A100s without early stopping. All models were compiled~\cite{Tan2026-fm} and accelerated using 
OpenEquivariance and cuEquivariance tensor-product kernels~\cite{Bharadwaj2025-ta}. 

\begin{figure*}[!p]
  \vskip 0.2in
  \begin{center}
    \centerline{\includegraphics[width=\columnwidth]{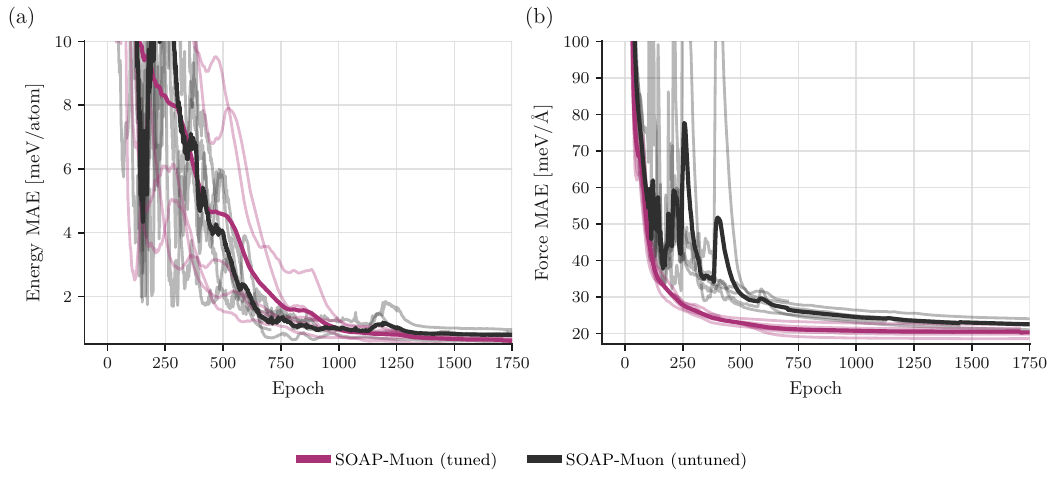}}
    \caption{Force and per-atom energy validation training curves for a tuned SOAP-Muon ($\rho=0.5$, $(\beta_1,\beta_2)=(0.95,0.95)$ for both parameter groups) and an untuned SOAP-Muon ($\rho=0.0$, $(\beta_1,\beta_2)=(0.95,0.98)$ for the Muon parameter group and $(\beta_1,\beta_2)=(0.9,0.95)$ for the Adam parameter group). Solid lines are per-epoch means, while the faint background lines show each individual seed run. Both tuned and untuned variants use the same learning rate ($10^{-3}$) and weight decay ($10^{-4}$).
    }
    \label{fig:H2Otune}
  \end{center}
\end{figure*}

\paragraph{SOAP-Muon Stability}
For the two model--system configurations considered in this work, we observed different sensitivities to the SOAP-Muon hyperparameters: the momentum parameters $(\beta_1,\beta_2)$ for each parameter group and the singular-value power $\rho$ applied to the singular-value matrix. Allegro-CDP was relatively robust to these choices, whereas NequIP-water exhibited substantially greater sensitivity and benefited more from further tuning (Figure~\ref{fig:H2Otune}).
For the momentum parameters, we searched over
$(\beta_1,\beta_2)\in\{(0.9,0.95), (0.95,0.95), (0.95,0.98)\}$,
following the combination of values reported in \citet{jordan2024muon,vyas2025soapimprovingstabilizingshampoo,vyas2025improving}. For the singular-value power, we considered $\rho\in\{0,0.5\}$ following \citet{vyas2025improving}. We used $\rho=0$ whenever performance differences were small, in order to avoid the additional cost of the full SVD computation.
For Allegro-CDP, we used $(\beta_1,\beta_2)=(0.95,0.98)$ for the Muon parameter group, $(0.9,0.95)$ for the Adam parameter group, and $\rho=0$. For NequIP-water, we used $(\beta_1,\beta_2)=(0.95,0.95)$ for both the Muon and Adam parameter groups, together with $\rho=0.5$. Since NequIP-water was more sensitive to these hyperparameters, we treat those settings as a practical default. However, broader evaluations across architectures and datasets are needed to establish more general hyperparameter recommendations. The dataset-specific instability observed for SOAP-Muon is consistent with \citet{vyas2025improving}, who found that optimizer stability varied across datasets: some datasets were stable without modification, whereas others required stability-enhancing adjustments. They further showed that datasets stable under the original optimizer were relatively insensitive to these modifications.
Nevertheless, our conclusion is based on the limited set of experiments performed here. 
We note that SOAP-Muon required substantially smaller learning rates than the other optimizers. Learning rates that were optimal for the other optimizers often led to severe training instabilities when used with SOAP-Muon. Throughout our experiments, the best-performing SOAP-Muon learning rate was approximately one order of magnitude lower than those used for the other optimizers (Table~\ref{tab:hyperparameters}).




\begin{table*}[htbp]
  \caption{Hyperparameter settings for each task, force-\% setting, and optimizer.}
  \label{tab:hyperparameters}
  \begin{center}
    \begin{small}
      \begin{sc}
        \begin{tabular}{lllcccc}
          \toprule
          \multirow{2}{*}{Task} & \multirow{2}{*}{Force \%} & \multirow{2}{*}{Optimizer}
            & \multicolumn{2}{c}{CDP (Allegro)}
            & \multicolumn{2}{c}{Water (NequIP)} \\
          \cmidrule(lr){4-5} \cmidrule(lr){6-7}
            & & & Learning rate & Weight decay & Learning rate & Weight decay \\
          \midrule
          \multirow{20}{*}{E+F}
            & \multirow{4}{*}{100}
            & AdamW     & $3{\times}10^{-2}$ & $1{\times}10^{-4}$ & $1{\times}10^{-2}$ & $1{\times}10^{-4}$ \\
            & & Muon      & $3{\times}10^{-2}$ & $1{\times}10^{-3}$ & $1{\times}10^{-2}$ & $1{\times}10^{-5}$ \\
            & & SOAP      & $3{\times}10^{-2}$ & $1{\times}10^{-4}$ & $1{\times}10^{-2}$ & $1{\times}10^{-4}$ \\
            & & SOAP-Muon & $1{\times}10^{-3}$ & $1{\times}10^{-5}$ & $1{\times}10^{-3}$ & $1{\times}10^{-4}$ \\
          \cmidrule(lr){2-7}
            & \multirow{4}{*}{75}
            & AdamW     & $3{\times}10^{-2}$ & $1{\times}10^{-3}$ & $1{\times}10^{-2}$ & $0$ \\
            & & Muon      & $3{\times}10^{-2}$ & $1{\times}10^{-3}$ & $1{\times}10^{-2}$ & $1{\times}10^{-4}$ \\
            & & SOAP      & $3{\times}10^{-2}$ & $1{\times}10^{-4}$ & $5{\times}10^{-3}$ & $0$ \\
            & & SOAP-Muon & $1{\times}10^{-3}$ & $0$                & $1{\times}10^{-3}$ & $0$ \\
          \cmidrule(lr){2-7}
            & \multirow{4}{*}{50}
            & AdamW     & $3{\times}10^{-2}$ & $1{\times}10^{-3}$ & $1{\times}10^{-2}$ & $0$ \\
            & & Muon      & $3{\times}10^{-2}$ & $1{\times}10^{-3}$ & $5{\times}10^{-3}$ & $0$ \\
            & & SOAP      & $3{\times}10^{-2}$ & $1{\times}10^{-4}$ & $1{\times}10^{-2}$ & $0$ \\
            & & SOAP-Muon & $1{\times}10^{-3}$ & $0$                & $1{\times}10^{-3}$ & $1{\times}10^{-4}$ \\
          \cmidrule(lr){2-7}
            & \multirow{4}{*}{10}
            & AdamW     & $1{\times}10^{-2}$ & $1{\times}10^{-3}$ & $1{\times}10^{-2}$ & $1{\times}10^{-5}$ \\
            & & Muon      & $1{\times}10^{-2}$ & $1{\times}10^{-3}$ & $1{\times}10^{-2}$ & $1{\times}10^{-4}$ \\
            & & SOAP      & $1{\times}10^{-2}$ & $1{\times}10^{-3}$ & $1{\times}10^{-2}$ & $1{\times}10^{-4}$ \\
            & & SOAP-Muon & $1{\times}10^{-3}$ & $1{\times}10^{-4}$ & $1{\times}10^{-3}$ & $0$ \\
          \cmidrule(lr){2-7}
            & \multirow{4}{*}{5}
            & AdamW     & $1{\times}10^{-2}$ & $1{\times}10^{-3}$ & $5{\times}10^{-3}$ & $1{\times}10^{-4}$ \\
            & & Muon      & $1{\times}10^{-2}$ & $1{\times}10^{-3}$ & $1{\times}10^{-2}$ & $1{\times}10^{-4}$ \\
            & & SOAP      & $1{\times}10^{-2}$ & $1{\times}10^{-5}$ & $5{\times}10^{-3}$ & $1{\times}10^{-3}$ \\
            & & SOAP-Muon & $1{\times}10^{-3}$ & $1{\times}10^{-4}$ & $1{\times}10^{-3}$ & $1{\times}10^{-5}$ \\
          \midrule
          \multirow{4}{*}{E}
            & \multirow{4}{*}{0}
            & AdamW     & $1{\times}10^{-2}$ & $0$                & $5{\times}10^{-3}$ & $1{\times}10^{-3}$ \\
            & & Muon      & $3{\times}10^{-3}$ & $1{\times}10^{-5}$ & $5{\times}10^{-3}$ & $0$ \\
            & & SOAP      & $3{\times}10^{-3}$ & $1{\times}10^{-5}$ & $5{\times}10^{-3}$ & $1{\times}10^{-4}$ \\
            & & SOAP-Muon & $3{\times}10^{-4}$ & $0$                & $3{\times}10^{-4}$ & $1{\times}10^{-3}$ \\
          \bottomrule
        \end{tabular}
      \end{sc}
    \end{small}
  \end{center}
\end{table*}

\section{Full Results} \label{app:full}
\begin{table*}[t]
  \caption{Test MAE aggregated across runs (mean $\scriptstyle\pm$ std) across 5 seeds. Gray force entries indicate metrics for targets not included in the task focus. \textbf{Bold} and \underline{underlined} entries mark the best and second-best mean values within each force-\% setting and metric, respectively.}
  \label{tab:test_metrics_cdp_h2o_mae}
  \begin{center}
    \begin{small}
      \begin{sc}
        \begin{tabular}{lllcccc}
          \toprule
          \multirow{2}{*}{Task} & \multirow{2}{*}{Force \%} & \multirow{2}{*}{Optimizer}
            & \multicolumn{2}{c}{CDP (Allegro)} & \multicolumn{2}{c}{Water (NequIP)} \\
          \cmidrule(lr){4-5} \cmidrule(lr){6-7}
            & & & E [$\mathrm{meV/atom}$] & F [$\mathrm{meV/\AA}$] & E [$\mathrm{meV/atom}$] & F [$\mathrm{meV/\AA}$] \\
          \midrule
          \multirow{20}{*}{E+F}
            & \multirow{4}{*}{100}
            & AdamW     & $0.628 \scriptstyle\pm 0.0434$                    & $32.2 \scriptstyle\pm 0.615$               & $0.773 \scriptstyle\pm 0.0713$                    & $25.7 \scriptstyle\pm 1.44$ \\
            & & Muon      & \underline{$0.581 \scriptstyle\pm 0.0328$}        & $29.6 \scriptstyle\pm 0.617$               & $1.53 \scriptstyle\pm 0.317$                     & $26.6 \scriptstyle\pm 2.90$ \\
            & & SOAP      & $\mathbf{0.569 \scriptstyle\pm 0.0694}$           & \underline{$29.6 \scriptstyle\pm 0.305$}   & \underline{$0.604 \scriptstyle\pm 0.0105$}        & $\mathbf{20.9 \scriptstyle\pm 0.698}$ \\
            & & SOAP-Muon & $0.582 \scriptstyle\pm 0.0469$                    & $\mathbf{27.8 \scriptstyle\pm 0.634}$      & $\mathbf{0.590 \scriptstyle\pm 0.0687}$           & \underline{$21.0 \scriptstyle\pm 1.04$} \\
          \cmidrule(lr){2-7}
            & \multirow{4}{*}{75}
            & AdamW     & $0.664 \scriptstyle\pm 0.0383$                    & $33.9 \scriptstyle\pm 0.993$               & $0.738 \scriptstyle\pm 0.117$                     & $25.1 \scriptstyle\pm 1.34$ \\
            & & Muon      & \underline{$0.622 \scriptstyle\pm 0.0287$}        & \underline{$31.1 \scriptstyle\pm 1.11$}    & $0.773 \scriptstyle\pm 0.146$                     & $27.1 \scriptstyle\pm 1.76$ \\
            & & SOAP      & $0.630 \scriptstyle\pm 0.0185$                    & $31.3 \scriptstyle\pm 0.617$               & $\mathbf{0.601 \scriptstyle\pm 0.0381}$           & $\mathbf{20.9 \scriptstyle\pm 0.872}$ \\
            & & SOAP-Muon & $\mathbf{0.596 \scriptstyle\pm 0.0280}$           & $\mathbf{30.1 \scriptstyle\pm 0.452}$      & \underline{$0.602 \scriptstyle\pm 0.101$}         & \underline{$21.6 \scriptstyle\pm 1.27$} \\
          \cmidrule(lr){2-7}
            & \multirow{4}{*}{50}
            & AdamW     & $0.698 \scriptstyle\pm 0.0175$                    & $37.4 \scriptstyle\pm 0.778$               & $0.789 \scriptstyle\pm 0.147$                     & $26.8 \scriptstyle\pm 1.30$ \\
            & & Muon      & \underline{$0.645 \scriptstyle\pm 0.0508$}        & \underline{$34.1 \scriptstyle\pm 0.803$}   & $1.01 \scriptstyle\pm 0.153$                      & $30.8 \scriptstyle\pm 2.86$ \\
            & & SOAP      & $0.690 \scriptstyle\pm 0.0724$                    & $34.5 \scriptstyle\pm 0.869$               & $\mathbf{0.650 \scriptstyle\pm 0.0772}$           & $\mathbf{22.7 \scriptstyle\pm 1.01}$ \\
            & & SOAP-Muon & $\mathbf{0.637 \scriptstyle\pm 0.00589}$          & $\mathbf{32.5 \scriptstyle\pm 1.01}$       & \underline{$0.713 \scriptstyle\pm 0.191$}         & \underline{$23.1 \scriptstyle\pm 1.07$} \\
          \cmidrule(lr){2-7}
            & \multirow{4}{*}{10}
            & AdamW     & $1.28 \scriptstyle\pm 0.0809$                     & $68.4 \scriptstyle\pm 2.54$                & $1.09 \scriptstyle\pm 0.198$                      & \underline{$34.5 \scriptstyle\pm 3.69$} \\
            & & Muon      & $1.06 \scriptstyle\pm 0.106$                      & $60.0 \scriptstyle\pm 3.10$                & $1.15 \scriptstyle\pm 0.108$                      & $37.4 \scriptstyle\pm 2.17$ \\
            & & SOAP      & \underline{$1.02 \scriptstyle\pm 0.0579$}         & \underline{$52.9 \scriptstyle\pm 1.65$}    & \underline{$1.04 \scriptstyle\pm 0.207$}          & $\mathbf{31.9 \scriptstyle\pm 2.19}$ \\
            & & SOAP-Muon & $\mathbf{0.977 \scriptstyle\pm 0.0525}$           & $\mathbf{51.9 \scriptstyle\pm 0.842}$      & $\mathbf{0.999 \scriptstyle\pm 0.133}$            & $40.6 \scriptstyle\pm 17.3$ \\
          \cmidrule(lr){2-7}
            & \multirow{4}{*}{5}
            & AdamW     & $1.62 \scriptstyle\pm 0.0551$                     & $94.9 \scriptstyle\pm 5.09$                & \underline{$1.51 \scriptstyle\pm 0.224$}          & \underline{$41.3 \scriptstyle\pm 2.09$} \\
            & & Muon      & $1.38 \scriptstyle\pm 0.103$                      & $84.0 \scriptstyle\pm 6.70$                & $\mathbf{1.46 \scriptstyle\pm 0.253}$             & $43.5 \scriptstyle\pm 1.10$ \\
            & & SOAP      & $\mathbf{1.16 \scriptstyle\pm 0.0841}$            & \underline{$69.2 \scriptstyle\pm 1.83$}    & $1.62 \scriptstyle\pm 0.659$                      & $41.9 \scriptstyle\pm 1.59$ \\
            & & SOAP-Muon & \underline{$1.20 \scriptstyle\pm 0.0901$}         & $\mathbf{68.1 \scriptstyle\pm 3.05}$       & $1.54 \scriptstyle\pm 0.466$                      & $\mathbf{41.2 \scriptstyle\pm 3.31}$ \\
          \midrule
          \multirow{4}{*}{E}
            & \multirow{4}{*}{0}
            & AdamW     & $5.16 \scriptstyle\pm 0.178$                      & \g{$503 \scriptstyle\pm 31.8$}             & $3.21 \scriptstyle\pm 0.272$                      & \g{$306 \scriptstyle\pm 46.1$} \\
            & & Muon      & $3.30 \scriptstyle\pm 0.135$                      & \g{$281 \scriptstyle\pm 9.73$}             & $5.37 \scriptstyle\pm 0.813$                      & \g{$591 \scriptstyle\pm 59.1$} \\
            & & SOAP      & \underline{$3.03 \scriptstyle\pm 0.389$}          & \g{\underline{$214 \scriptstyle\pm 25.7$}} & $\mathbf{2.38 \scriptstyle\pm 0.573}$             & \g{$\mathbf{236 \scriptstyle\pm 54.8}$} \\
            & & SOAP-Muon & $\mathbf{2.75 \scriptstyle\pm 0.163}$             & \g{$\mathbf{201 \scriptstyle\pm 18.8}$}    & \underline{$2.77 \scriptstyle\pm 0.346$}          & \g{\underline{$269 \scriptstyle\pm 24.5$}} \\
          \bottomrule
        \end{tabular}
      \end{sc}
    \end{small}
  \end{center}
\end{table*}


\begin{figure*}[ht]
  \vskip 0.2in
  \begin{center}
    \centerline{\includegraphics[width=\columnwidth]{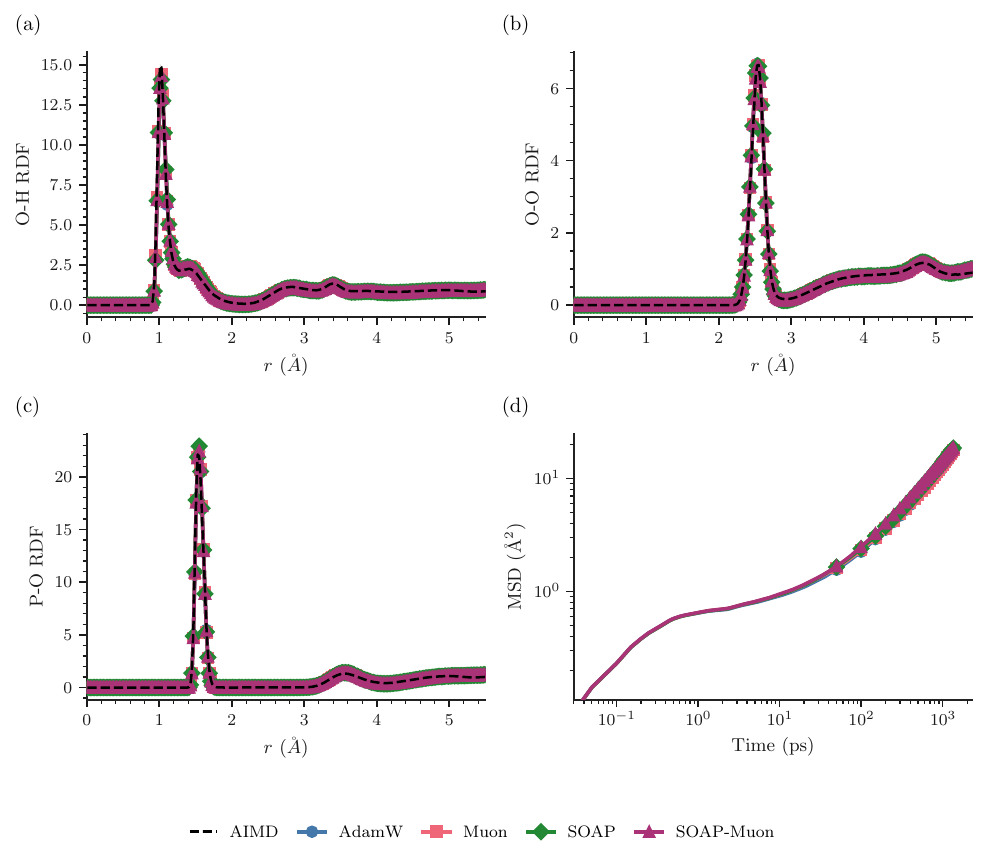}}
    \caption{Physical observables obtained from MD simulations of CDP using energy and force trained MLIPs with AdamW (blue), Muon (orange), SOAP (green), and SOAP-Muon (purple), compared against ab initio reference curves. (a–c) Radial distribution functions for the O–H, O–O, and P–O pairs, respectively. (d) Mean squared displacement of H calculated from the corresponding MD trajectories.
    }
    \label{fig:CDPfid}
  \end{center}
\end{figure*}

\begin{figure*}[ht]
  \vskip 0.2in
  \begin{center}
    \centerline{\includegraphics[width=\columnwidth]{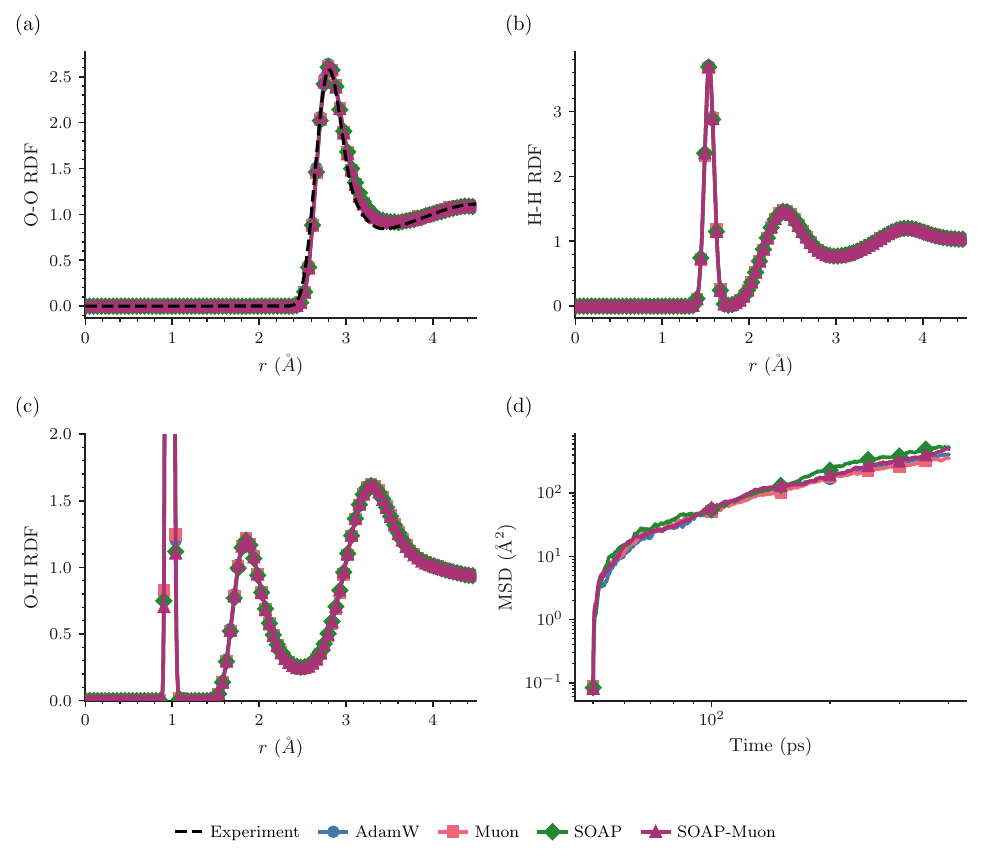}}
    \caption{Physical observables obtained from MD simulations of water using energy-and-force-trained MLIPs with AdamW (blue), Muon (orange), SOAP (green), and SOAP-Muon (purple). (a–c) Radial distribution functions (RDFs) for the O–O, H–H, and O–H pairs, respectively. O-O RDF is compared against experimental data from \citet{Skinner2014Structure} (d) Mean squared displacement calculated from the corresponding MD trajectories.
    }
    \label{fig:H2Ofid}
  \end{center}
\end{figure*}

\begin{figure*}[ht]
  \vskip 0.2in
  \begin{center}
    \centerline{\includegraphics[width=\columnwidth]{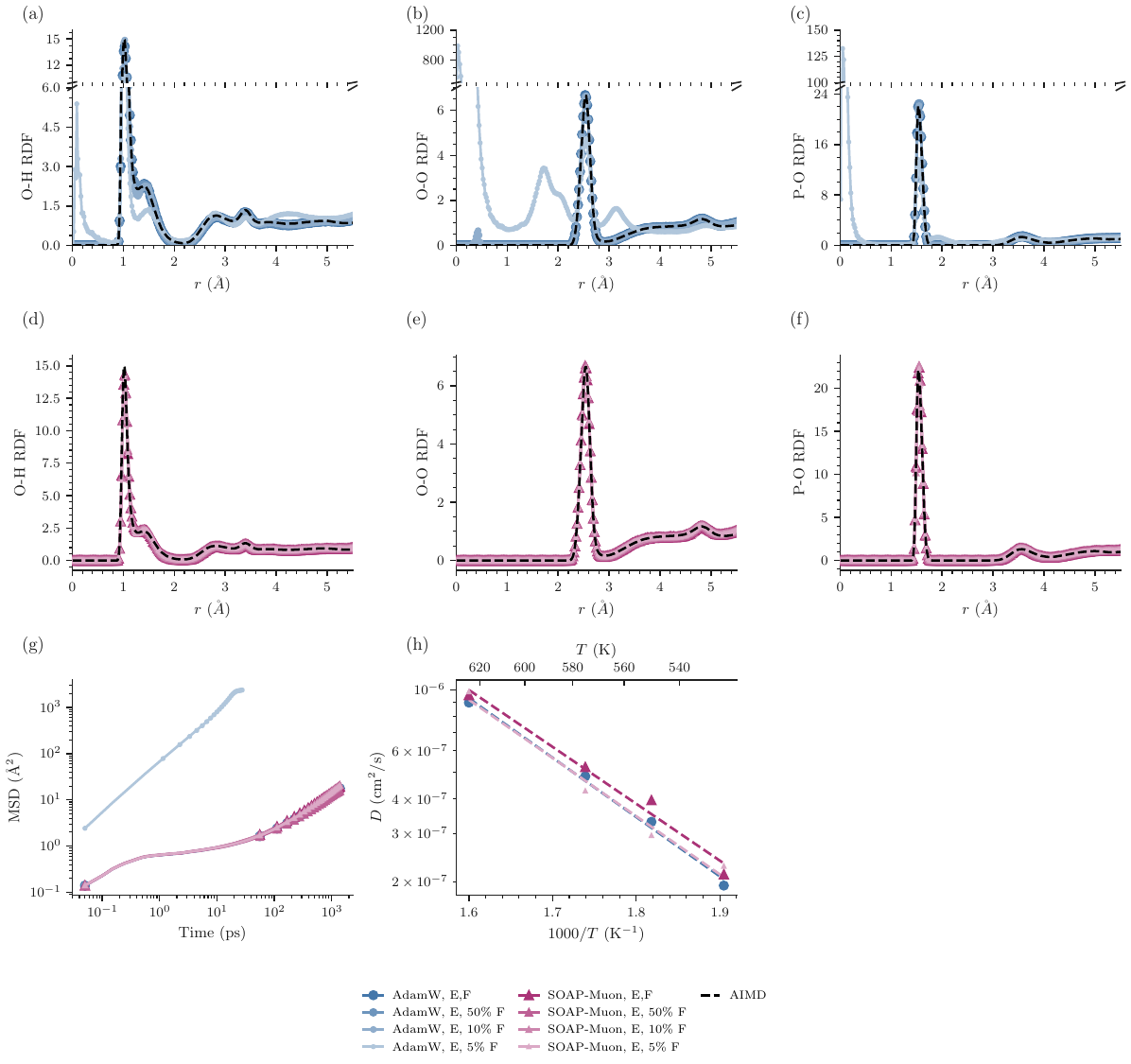}}
    \caption{Physical observables obtained from MD simulations of CDP using MLIPs trained with AdamW (blue) and SOAP-Muon (purple), on energy labels and varying fractions of force labels: $5\%$, $10\%$, $50\%$, and $100\%$ force supervision (lighter to darker shades indicate increasing fractions of force labels). (a-c) RDFs for the O–H, O–O, and P–O pairs with AdamW. (d-f) RDFs for the O–H, O–O, and P–O pairs with SOAP-Muon. (g) MSD of H calculated from the corresponding MD trajectories. (h) Arrhenius plot of the diffusion coefficient $D$, derived from the linear regime of the proton MSD, versus $1/T$. The activation energy is determined from the slope of the fitted linear regression --- AdamW (E, F): $E_a=0.429\,\mathrm{eV}$, SOAP-Muon (E, F): $E_a=0.411\,\mathrm{eV}$, SOAP-Muon (E, $5\%$ F): $E_a=0.421\,\mathrm{eV}$. MD simulations with the AdamW (E, $5\%$ F) model did not remain stable long enough to reach the linear proton-diffusion regime. The experimental range is $E_a = 0.39\text{--}0.43\;\mathrm{eV}$ \cite{Haile2007Solid,Ishikawa2008Proton,wang2025revealingprotonslingshotmechanism}.
    }
    \label{fig:CDPsparse}
  \end{center}
\end{figure*}

\begin{figure*}[ht]
  \vskip 0.2in
  \begin{center}
    \centerline{\includegraphics[width=\columnwidth]{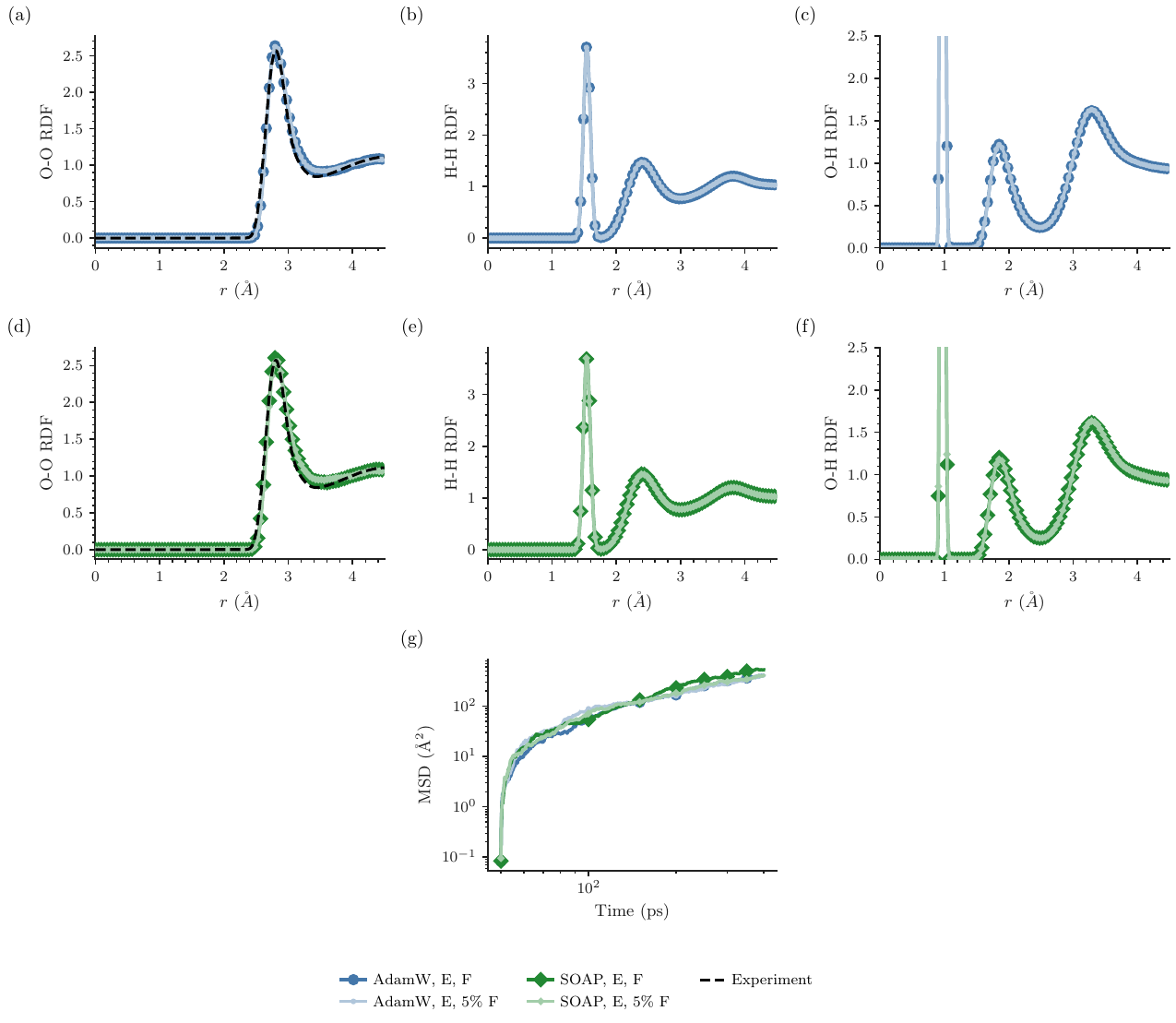}}
    \caption{Physical observables obtained from MD simulations of water using MLIPs trained with AdamW (blue) and SOAP (green), on energy labels and varying fractions of force labels: $5\%$ and $100\%$ force supervision (lighter to darker shades indicate increasing fractions of force labels). (a-c) RDFs for the O--O, O--H, and H--H pairs with AdamW. (d-f) RDFs for the O--O, O--H, and H--H pairs with SOAP-Muon. (g) MSD calculated from the corresponding MD trajectories.
}
    \label{fig:H2Osparse}
  \end{center}
\end{figure*}


\end{document}